\documentclass{article}

\usepackage{microtype}
\usepackage{graphicx}
\usepackage{booktabs} 
\usepackage{amsmath}
\usepackage{graphicx}
\usepackage{subcaption}

\usepackage{hyperref}

\usepackage[accepted]{icml2019}

\icmltitlerunning{Affine Variational Autoencoders}

\begin{document}

\twocolumn[
\icmltitle{Affine Variational Autoencoders: An Efficient Approach for Improving Generalization and Robustness to Distribution Shift}




\begin{icmlauthorlist}
\icmlauthor{Rene Bidart}{waterloo,waii}
\icmlauthor{Alexander Wong}{waterloo,waii}
\end{icmlauthorlist}

\icmlaffiliation{waterloo}{Systems Design Engineering, University of Waterloo, Waterloo, ON, Canada }
\icmlaffiliation{waii}{Waterloo Artificial Intelligence Institute, Waterloo, ON, Canada}

\icmlcorrespondingauthor{Rene Bidart}{rbbidart@uwaterloo.ca}

\icmlkeywords{Machine Learning, ICML}

\vskip 0.3in
]



\printAffiliationsAndNotice{}  

\begin{abstract}
In this study, we propose the \textit{Affine Variational Autoencoder (AVAE)}, a variant of Variational Autoencoder (VAE) designed to improve robustness by overcoming the inability of VAEs to generalize to distributional shifts in the form of affine perturbations. By optimizing an affine transform to maximize ELBO, the proposed AVAE transforms an input to the training distribution without the need to increase model complexity to model the full distribution of affine transforms. In addition, we introduce a training procedure to create an efficient model by learning a subset of the training distribution, and using the AVAE to improve generalization and robustness to distributional shift at test time. Experiments on affine perturbations demonstrate that the proposed AVAE significantly improves generalization and robustness to distributional shift in the form of affine perturbations without an increase in model complexity.

\end{abstract}
\vspace{-0.3in}
\section{Introduction}
\label{submission}
\begin{figure}
  \centering
    \includegraphics[width=.25\textwidth]{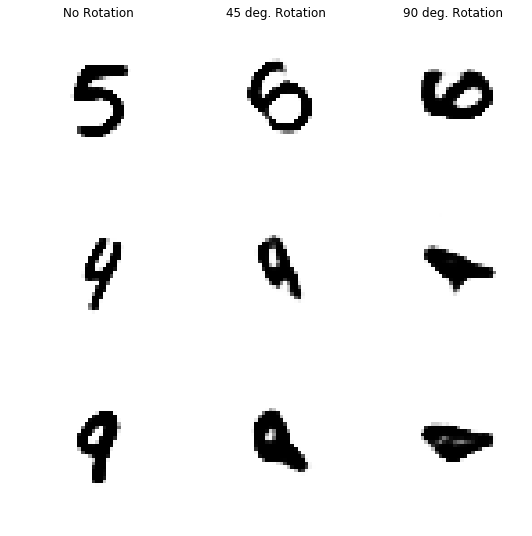}
    \vspace{-0.22in}
    \caption{Examples of reconstructed images after various rotational perturbations from a VAE trained on MNIST. The quality of the reconstructed images degrade significantly under increasing rotational perturbation.}
    \label{fig:vae_loss_rotation_ex}
\vspace{-0.26in}
\end{figure}

\vspace{-0.1in}
While deep neural networks have been shown to be extremely powerful, they are often quite fragile and do not generalize well to distributional shifts without explicit methods to account for such shifts. A particular form of distributional shift that has seen recent attention are affine data perturbations, where out-of-distribution samples can be treated as affine transformed variants of samples within the training distribution. As an example, Fig.~\ref{fig:vae_loss_rotation_ex} shows examples of the inability of a variational autoencoder (VAE) to encode and decode images after various rotational perturbations.

Existing methods for improving robustness to affine perturbation have limitations, including data augmentation which forces an increase in model complexity (Section 4.1), explicit equivariance which results in an increase in memory use and is difficult to scale to more transforms\cite{cohen2016group}, or are approximate methods, where the model must estimate a transformation to bring the image back to a canonical orientation~\cite{stn}.

In this study, we introduce the \textit{Affine Variational Autoencoder (AVAE)}, an extension of the Variational Autoencoder (VAE) designed to improve generalization and robustness to distribution shift related to affine perturbation. We take an alternative approach by explicitly leveraging VAE's loss to evaluate if a sample is within the training distribution, with the corresponding affine transform performed within the AVAE optimized to reduce this loss. Instead of increasing model complexity as used in data augmentation, we increase computational complexity to transform a sample back within the distribution the model was trained on, with the added benefit of returning the affine transform needed to bring the sample back to the training distribution.

More specifically, we introduce a method for enabling pre-trained VAEs, which follow a certain training distribution, to generalize to the full set of affine transforms, and a way to train a model on the full distribution, without increasing the model capacity. Given some data $X$, we consider the full set of possible affine transforms of $X$ to be $X_{full}$, and a subset of these transforms as $X_{sub}$. These methods are:
\vspace{-0.2in}
\begin{enumerate}
    \item A module added to a pre-trained VAE trained on $X_{sub}$, allowing for improved generalization to samples from $X_{full}$ through optimization of an affine transform
    \vspace{-0.05in}
    \item A training procedure to allow the AVAE to be trained on the full dataset, $X_{full}$, by optimizing the affine transform for each batch during training, allowing a model to be trained on the full dataset without forcing an increase in model capacity.
\end{enumerate}

\vspace{-0.1in}

\begin{figure}
  \centering
    \includegraphics[width=.5\textwidth]{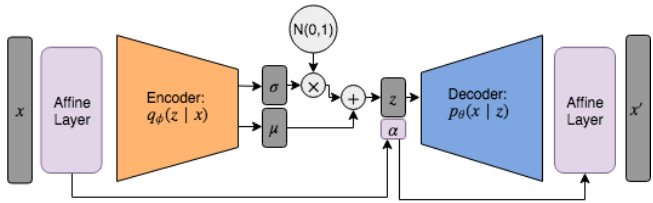}
    \vspace{-0.2in}
    \caption{Affine Variational Autoencoder (AVAE).  This extends upon the conventional VAE by introducing two affine layers. The first affine layer performs a learned affine transform to the input, parameterized by $\alpha$. This is encoded and decoded by the encoder and decoder, respectively, and finally the second affine layer performs the inverse transform, producing the final output, $x'$.}
    \label{fig:affine_vae_1d}
\vspace{-0.2in}
\end{figure}

\vspace{-0.1in}
\section{Background and Related Work}
\vspace{-0.1in}
\subsection{Variational autoencoders}
\vspace{-0.1in}
Variational autoencoders~\cite{vae} are generative models where it is assumed that the data, $X=\{x\}_{i=1}^n$ are generated from latent variables. An encoder and decoder network are simultaneously trained, where the encoder models the posterior distribution of latent variables $z$, given an image $x$ as $q_\phi(z|x)$, and a decoder that gives the distribution of data (in this case, an image) conditioned this latent variable. It is assumed that the latent variables are independent standard normal.

\vspace{-0.3in}
\begin{equation} \label{objective}
-L_{VAE} = E_{z\sim q_\phi}[logp_\rho(x|z)] - KL(q_\phi(z|x) || p_\rho(z))]
\end{equation}

The objective of the VAE is to maximize the evidence lower bound (ELBO) shown in equation~\ref{objective}. In many cases, convolutional neural networks (CNNs) are used to approximate both $p$ and $q$. This entire network is differentiable, so can be trained using stochastic gradient descent with this loss.
\vspace{-0.1in}
\subsection{Generalization to Affine Transforms}
\vspace{-0.1in}
There have been many attempts to both learn representations that are robust to distributional shifts under a set of transformations as well as increasing interpretability by encoding these transforms explicitly in the latent space for uses such as image classification, compression, and generation.

\textbf{Spatial Transformer Networks} (STN)~\cite{stn} apply an affine transform to the input image, transforming it to some canonical orientation. This can be described in terms of three components:
\vspace{-0.1in}
\begin{enumerate}
    \item \textbf{Localization Network}: This is a neural network taking the input image and outputting the affine transformation parameters, $\alpha \in {\rm I\!R}^6$ to be applied.
    \vspace{-0.1in}
    \item \textbf{Sampling Grid}: Given an affine transform, the grid of coordinates in the input associated with each point in the output
    \vspace{-0.1in}
    \item \textbf{Data sampling}: Given the grid, use bilinear sampling to apply it to the input.
\end{enumerate}
\vspace{-0.1in}
More generally, there has been research on learning disentangled representations where semantically relevant variables are explicit in the latent space~\cite{ridgeway2016survey}. In general these are not limited to affine transforms, and include variations such a lighting, color, or physical attributes like shape. One approach is based on semi-supervised learning, where images are generated based on both a latent variable and some relevant factor of variation, which are assumed to be independent~\cite{kingma2014semi}. For face generation, disentangling shape and appearance was tackled through the synthesis of appearance on a template followed by a deformation~\cite{shu2018deforming}. Other work divides the latent space into explicit and implicit factors of variation, and training process of varying only one factor while fixing the others is used to enforce the disentangled latent space~\cite{kulkarni2015deep}. These methods all require supervised inputs, where they are labeled based on some factor of variation. Other work has created networks that are equivariant to one specific factor of variation, for example, constructing deep convolutional neural networks that are equivariant to rotation and reflection~\cite{cohen2016group}. While this is an interesting method, adding more factors of variation in this way increases the complexity dramatically, so is difficult to scale.
\vspace{-0.1in}

\section{Affine Variational Autoencoders}
\vspace{-0.1in}
The proposed notion of affine variational autoencoders (AVAEs) can be described as follows.  For a given input $x$, we can estimate how far it is outside the training distribution, $X_{sub}$ using the ELBO, as shown in Equation~\ref{objective}. In this work, we focus on out-of-distribution samples that can be treated as affine transform variants of in-distribution samples.  Therefore, given a sample $x\notin X_{sub}$, we formulate the problem as transforming $x$ to $x_{A}\in X_{sub}$ by optimizing the affine transform to reduce the VAE's loss.  This approach is useful because the model capacity required for a subset of the full distribution, $X_{sub}$ is less than that required for a model expected to generalize to the full distribution $X_{full}$. This procedure enables us to trade-off model complexity for computational complexity of performing this optimization.
\vspace{-0.05in}
\subsection{Model}
\vspace{-0.1in}
The AVAE extends upon the conventional VAE architecture with the introduction of two affine transform layers, before and after the conventional VAE, as shown in  Fig.~\ref{fig:affine_vae_1d}. An input sample is first fed into first affine layer, which performs an affine transform before passing it into the encoder for latent space representation.  The output of the decoder in the AVAE is fed into a second affine layer, which performs an inverse affine transform on the output of the decoder, producing the final output. The parameters of the affine transform applied to the input sample are appended to the latent space and later used in the final, inverse affine layer.  The affine layers are implemented similarly to the STN~\cite{stn} (see section 2.2), in the form of a sampling grid and data sampling, but instead of learning the affine transform parameters through a localization network, we optimize these parameters to reduce VAE loss.

\vspace{-0.05in}
\subsection{Optimization and Training}
\vspace{-0.1in}
Given an input sample $x$, the objective of the conventional VAE is to learn an encoder $q_\phi$, and decoder $p_\rho$ to maximize the likelihood of the data, by maximizing the ELBO as indicated in section 2.1. In the proposed AVAE, the addition of the affine layers allows us to optimize a transform, $A(x) = x_{A}$, to reduce the VAE loss of $A^{-1}(p_\rho(q_\rho(x_A)))$, maximizing the probability of this image being from the training distribution.  More specifically, in the AVAE, to encode a sample we must learn $\alpha$ that maximizes the ELBO in order to find the optimal affine transformation to apply to the input sample, which we can formulate as:
\vspace{-0.1in}
\begin{equation}
    \underset{\alpha}{\text{argmin}} \left\{L_{VAE}[A^{-1}(p_\rho(q_\phi(A(x))))]\right\}
    \vspace{-0.1in}
\end{equation}
\noindent where $A$ refers to the first affine layer, $A^{-1}$ refers to the second affine layer, both taking $\alpha$ as parameters. Note that VAE parameters $q_\phi, p_\rho$ remain unchanged during this.

This network is differentiable, so it is possible to take the derivative of the loss w.r.t. the affine transform parameters, $\alpha$, so this can be optimized using stochastic gradient descent. We find in practice there are issues with the optimizer being caught in local minima, so we use multiple random restarts, where we first try the loss at a set number of affine parameters, and only perform gradient descent on the best performing parameters.

Given a set of samples at some canonical orientation, such as MNIST, the AVAE can be trained identically to the conventional VAE, or the affine layers can be added to an existing pre-trained model. However, an issue with training an AVAE in an identical manner as a conventional VAE is that the benefits of generalization and robustness to distributional shift provided by the AVAE are diminished if the training data isn't already in a canonical orientation like in MNIST. Training on $X_{full}$ instead of $X_{sub}$ will require a greater capacity model, as indicted in section 4.1.
\vspace{-0.05in}
\subsection{Transformation Optimization During Training}
\vspace{-0.1in}
To preserve the benefits of the AVAE when the dataset spans the full set of possible affine transforms, we introduce an alternative training scheme where we force the model to learn to encode the images well at only a subset of the possible transformations, limiting the model capacity required. This enforcement to a subset of possible transformations is accomplished indirectly by iteratively optimizing the affine transform for each batch.

More specifically, we optimize the affine transform for each input sample before performing gradient descent on the encoder and decoder parameters during training.  This has the effect of transforming the training data to have similar orientational properties, reducing the number of possible transformations that the model must learn to encode and enabling one to reap the benefits of generalization and robustness to distributional shift provided by the AVAE.

\begin{figure}
        \vspace{-0.1in}
  \centering
    \includegraphics[width=.4\textwidth]{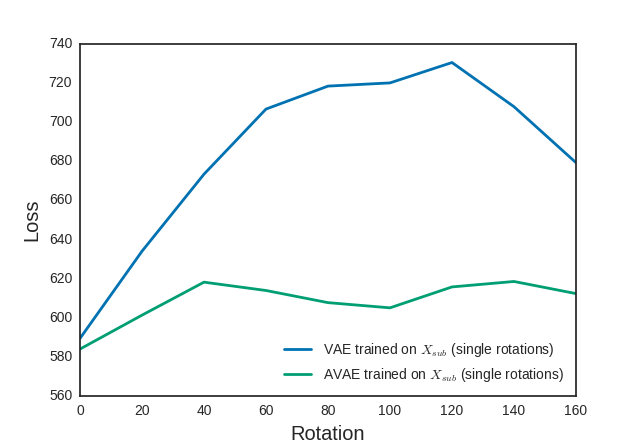}
        \vspace{-0.1in}
    \caption{Average loss of VAE and AVAE over the MNIST validation set under varying rotational perturbations. Both models were trained without data augmentation, so conventional VAE does not generalize well to new rotational perturbations as the input data deviates from the training set. The AVAE is robust to this distribution shift, as shown by the relatively flat loss across perturbations.}
    \vspace{-0.2in}
    \label{fig:vae_avae_loss_rotation}
\end{figure}

\vspace{-0.2in}
\section{Experiments}
\vspace{-0.1in}
To evaluate the efficacy of the proposed AVAE, we study its robustness to distributional shifts under various affine perturbation. We focus mostly on a specific type of affine perturbation in the form of rotational perturbation to study the behaviour of AVAEs in greater depth. Results for more general affine perturbations is also presented.
\begin{figure}
  \centering
    \includegraphics[width=.4\textwidth]{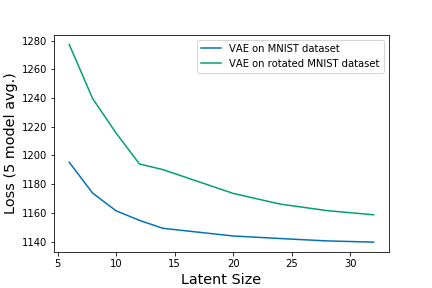}
    \vspace{-0.1in}
    \caption{Comparison of the loss associated with VAEs trained on standard MNIST dataset, and the dataset of all rotational perturbations. A larger latent size is required to generalize to the rotated dataset with comparable loss. This is loss averaged over 5 models.}
    \vspace{-0.3in}
    \label{fig:latent_size_perf}
\end{figure}
\vspace{-0.05in}
\subsection{Limitations of VAE and Data Augmentation}
\vspace{-0.1in}
VAEs do not generalize to out-of-distribution samples that are variants of in-distribution samples under rotational perturbations unless they are explicitly trained on this type of data. Fig.~\ref{fig:vae_avae_loss_rotation} shows the loss associated with a VAE encoding and decoding the MNIST validation set under different rotational perturbations. The performance of a conventional VAE decreases steadily as the rotational perturbations deviate more from the training set, with the loss reaching a maximum around $120^{\circ}$. It decreases after this because many digits are similar when rotated $180^{\circ}$.

The common approach to dealing with this limitation is data augmentation, directly training the model on $X_{full}$, but to train a VAE on a more diverse dataset requires a more complex model. We investigate this by altering the latent size in the model, as shown in Fig.~\ref{fig:latent_size_perf}. This could be partially avoided through the use of a more complex model instead of larger latent size, but reinforces the assertion that it is expensive to make a VAE generalize well to perturbations via data augmentation.

\vspace{-0.05in}
\subsection{AVAE on Rotational Perturbations}
\vspace{-0.1in}
We first investigate if a conventional VAE can be extended into a AVAE to generalize and be more robust to rotational perturbations. Fig.~\ref{fig:vae_avae_loss_rotation} shows the loss is reduced across all rotational perturbations by extending a conventional VAE into an AVAE. The increase in performance is greatest at the rotational perturbations that are furthest from the original training distribution, and the models perform similarly when rotational perturbations are close to the standard orientation the data was trained on, $0^{\circ}$.

\begin{figure}
  \centering
    \includegraphics[width=.4\textwidth]{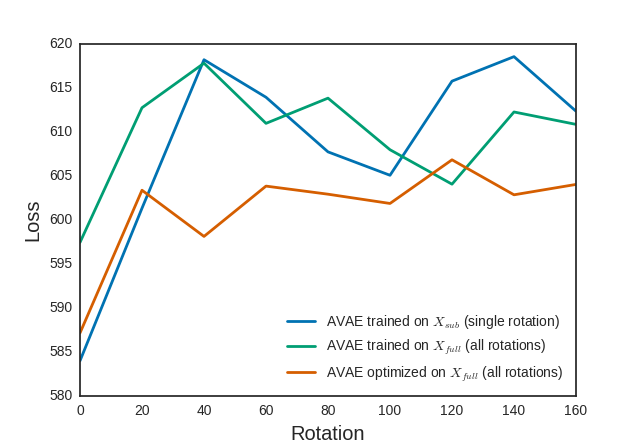}
    \vspace{-0.2in}
    \caption{Comparison of AVAE trained using MNIST with no rotation augmentation ($X_{sub}$), trained on rotation augmentations ($X_{full}$), and trained on rotation augmentations with the proposed transformation optimization during training. Training with transformation optimization generally outperforms the other approaches.}
    \label{fig:avae_loss_rotation}
    \vspace{-0.25in}
\end{figure}

\vspace{-0.2in}
\begin{table}[h!]
  \begin{center}
    \caption{{VAE and AVAE average validation loss over $0^{\circ}-180^{\circ}$ rotational perturbations on MNIST}}
    \label{tab:results}
    \begin{tabular}{l|c|r}
      \textbf{Model} & \textbf{Training Data} & \textbf{Avg. Loss}\\
      \hline
      AVAE (trans. opt.) & random perturbations & 601.2\\
      AVAE & single orientation & 608.6\\
      AVAE & random perturbations & 609.8\\
      VAE & single orientation & 684.4\\

    \end{tabular}
  \end{center}
  \vspace{-0.2in}
\end{table}

\subsection{Transformation Optimization During Training}
Earlier we speculated that optimizing the affine transformation training would allow the model to encode a smaller subset of the full distribution, enabling better performance for a given model complexity, with improved generalization to the rest of the distribution during test time. In appendix~1 we show that empirically this is the case, because as the model trains it learns to encode a progressively smaller portion of the original dataset.

In Table~\ref{tab:results}, we can see that over all rotational perturbations this seems to be true, with the AVAE using transformation optimization during training achieving the best results. The AVAE trained on a single orientation is slightly better than random perturbations because it isn't forced to encode the full dataset, but these models overall perform similarly.

Looking at Figure~\ref{fig:avae_loss_rotation}, these methods of training face a trade-off, because the rotation-augmented AVAE is better for encoding out-of-distribution samples, but the single orientation AVAE should perform better once the samples are brought into the correct orientation. Because the optimization process at test time isn't perfect, the rotation augmented AVAE still has some benefits for these out-of-distribution samples. The AVAE with transformation optimization during training effectively balances both these considerations, and based on Figure~\ref{fig:avae_loss_rotation} it is the superior model at almost all rotations.
\vspace{-0.05in}
\subsection{General Affine Transforms}
\begin{figure}
  \centering
    \includegraphics[width=.4\textwidth]{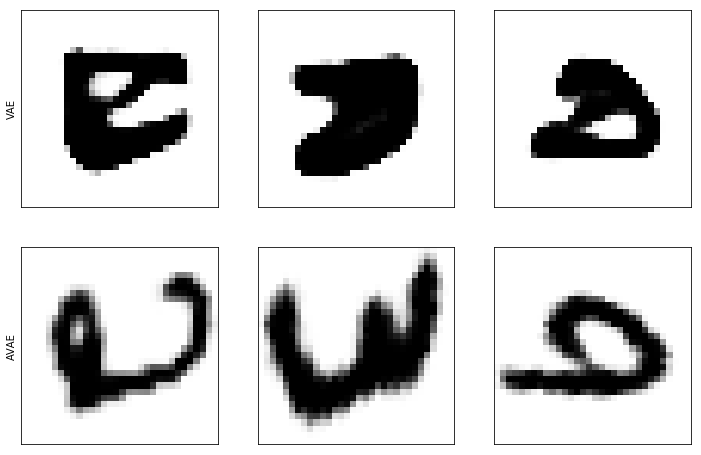}
    \vspace{-0.2in}
    \caption{Examples of the reconstructions of images transformed by the a random affine transformation from the VAE (top) and AVAE (bottom).}
    \label{fig:vae_avae_affine_ex}
    \vspace{-0.25in}
\end{figure}

\vspace{-0.1in}
We also investigated the generalization of AVAE to inputs perturbed by more general affine transforms through the use of the AVAE's optimization process. Inputs we perturbed by random rotations, random shears up to $55^{\circ}$, as well as random scaling by up to $50\%$. Based on these transforms, we found the AVAE was able to decrease the validation loss by $14\%$ compared to the conventional VAE. An example of this is shown in Figure~\ref{fig:vae_avae_affine_ex}.

\vspace{-0.1in}
\section{Conclusion}
\vspace{-0.1in}
In this study, we introduced affine variational autoencoders, extending upon VAEs to improve generalization and robustness to distributional shifts due to affine transformations by optimizing the affine transform based on the VAE loss.  In experiments using rotational and general affine perturbations, it was shown that the proposed AVAEs can indeed improve robustness in practice. This shows a practical trade off between optimization and model capacity, where to generalize to affine transforms we can substitute increased model capacity with this optimization procedure. In addition, we introduced a training procedure that forces the model to learn only a subset of the possible affine transformations, allowing the AVAE to be trained on any dataset.

\bibliography{bibfile}
\bibliographystyle{icml2019}

\appendix
\section{Appendix}
\vspace{-0.1in}
\subsection{Transformation Optimization During Training}
\begin{figure}
  \centering
    \includegraphics[width=.5\textwidth]{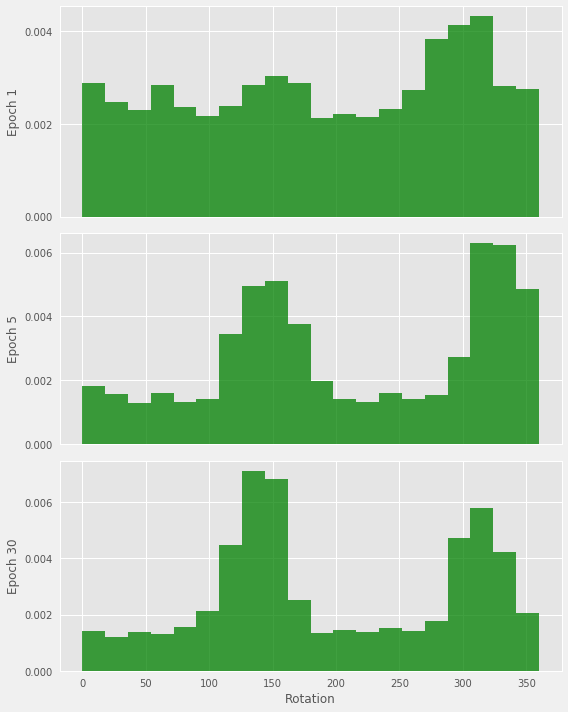}
    \vspace{-0.2in}
    \caption{Distribution of rotations of the "1" digit during training of the AVAE at epochs 1, 5 and 30. We can see as training progresses the AVAE learns to encode most digits at the same orientation as the most dense representation of the data.}
    \label{fig:rot_batch_opt_1}
    \vspace{-0.25in}
\end{figure}

\vspace{-0.1in}
Under the normal training process, the distribution of rotations the model is trained on should be uniformly distributed over $[0^{\circ}, 360^{\circ}]$. Through the optimization process during training, we find that the AVAE learns a more efficient representation where the digits of a given class are oriented at the same rotation. For the digit "1", the model learns to encode at two orientations $180^{\circ}$ apart because the "1" is almost identical when under this rotation, as shown in Figure~\ref{fig:rot_batch_opt_1}.

\begin{figure}
  \centering
    \includegraphics[width=.5\textwidth]{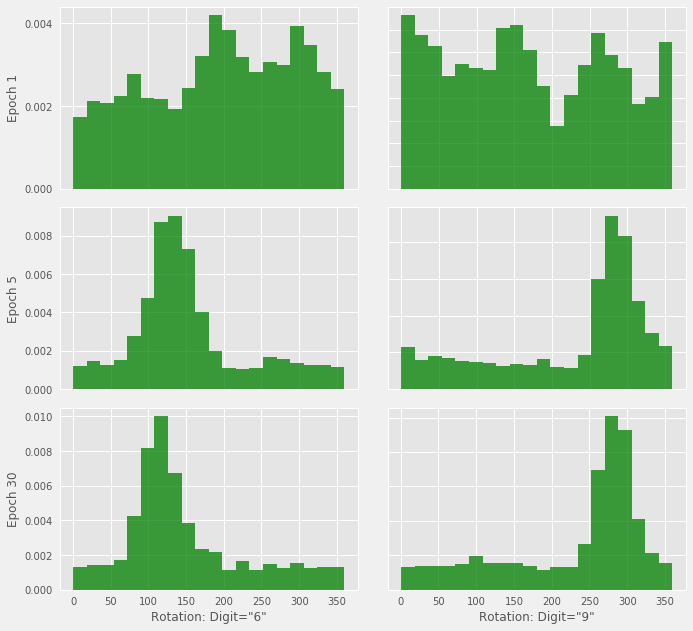}
    \vspace{-0.2in}
    \caption{Distribution of rotations of the "6" and "9" digits during training of the AVAE at epochs 1, 5 and 30. We can see as training progresses the AVAE learns to encode most digits at the same orientation, but additionally these numbers are encoded as $180^{\circ}$ rotations of one another.}
    \label{fig:rot_batch_opt_69}
    \vspace{-0.25in}
\end{figure}

The model learns to encode the "6" and "9" digits as $180^{\circ}$ rotations of one another, as this is the most compressed representation of the data. This is shown in Figure~\ref{fig:rot_batch_opt_69}.

\subsection{Implementation}
All experiments were implemented using Pytorch~\cite{paszke2017automatic}, and optimization was done using the Adam Optimizer~\cite{kingma2014adam} with the learning rate set to 0.001, weight decay set to 0.0005 and batch size of 256. The MNIST dataset was normalized by mean and standard deviation and 0 padded to 40x40 pixels.

We use a VAE architecture based on a previously successful implementation\cite{schott2018towards}, which uses and encoder composed of four convolutional layers of sizes [32, 32, 64, 16], and a decoder composed of transposed convolutions of sizes [32, 16, 16, 1]. For the first three layers of both the encoder and decoder, Exponential Linear Unit activation functions (ELU)\cite{DBLP:journals/corr/ClevertUH15} are used as well as batch normalization\cite{DBLP:journals/corr/IoffeS15}.

\end{document}